\documentclass{llncs}

\usepackage{graphicx}
\usepackage[pdftex, pagebackref=true, pdfpagelabels=true,
hypertexnames=true, plainpages=false, naturalnames=false,
bookmarks=true, bookmarksnumbered=true,
pdfpagemode=UseOutlines,]{hyperref}
\hypersetup{colorlinks, citecolor=[rgb]{0.0,0.0,0.5},
filecolor=[rgb]{0.6,0.0,0.0}, linkcolor=[rgb]{0.0,0.0,0.5},
urlcolor=[rgb]{0.0,0.0,0.5}}
\usepackage{url}            
\usepackage[numbers]{natbib}

\usepackage{graphicx} \graphicspath{ {figures/} }
\usepackage{subfigure}
\usepackage{tabularx}
\usepackage{booktabs}
\usepackage{multirow}
\usepackage{bbding}

\begin{document}
\title{Automatic Segmentation of Pulmonary Lobes Using a Progressive Dense V-Network}
\authorrunning{Abdullah-Al-Zubaer Imran et al.} 
%
\author{Abdullah-Al-Zubaer Imran\inst{1,2} \and Ali Hatamizadeh\inst{1,2} \and Shilpa P. Ananth\inst{2} \and \\Xiaowei Ding\inst{1,2,}\Envelope \and Demetri Terzopoulos\inst{1,2} \and Nima Tajbakhsh\inst{2}
}
\institute{University of California, Los Angeles, CA 90095, USA\\
\and
VoxelCloud Inc.,
Los Angeles, CA 90024, USA  \\
\email{xding@voxelcloud.io}
}

\maketitle              
\vspace{0.5cm}
\begin{abstract}
\vskip -1cm 
Reliable and automatic segmentation of lung lobes is important for diagnosis, assessment, and quantification of pulmonary diseases. The existing techniques are  prohibitively slow,  undesirably rely on prior (airway/vessel) segmentation, and/or require user interactions for optimal results. This work presents a reliable, fast, and fully automated lung lobe segmentation based on a progressive dense V-network (PDV-Net). The proposed method can segment lung lobes in one forward pass of the network, with an average runtime of 2 seconds using 1 Nvidia Titan XP GPU, eliminating the need for any prior atlases, lung segmentation or any subsequent user intervention. We evaluated our model using 
84 chest CT scans from the LIDC and 154 pathological cases from the LTRC datasets. Our model achieved a Dice score of $0.939 \pm 0.02$ for the LIDC test set and $0.950 \pm 0.01$ for the LTRC test set, significantly outperforming a 2D U-net model and a 3D dense V-net. We further evaluated our model against 55 cases from the LOLA11 challenge, obtaining an average Dice score of 0.935---a performance level competitive to the best performing team with an average score of 0.938. Our extensive robustness analyses also demonstrate that our model can reliably segment both healthy and pathological lung lobes in CT scans from different vendors, and that our model is robust against configurations of CT scan reconstruction.
\end{abstract}

\section{Introduction}
\begin{figure}[t]
\centering
\includegraphics[width=0.6\linewidth]{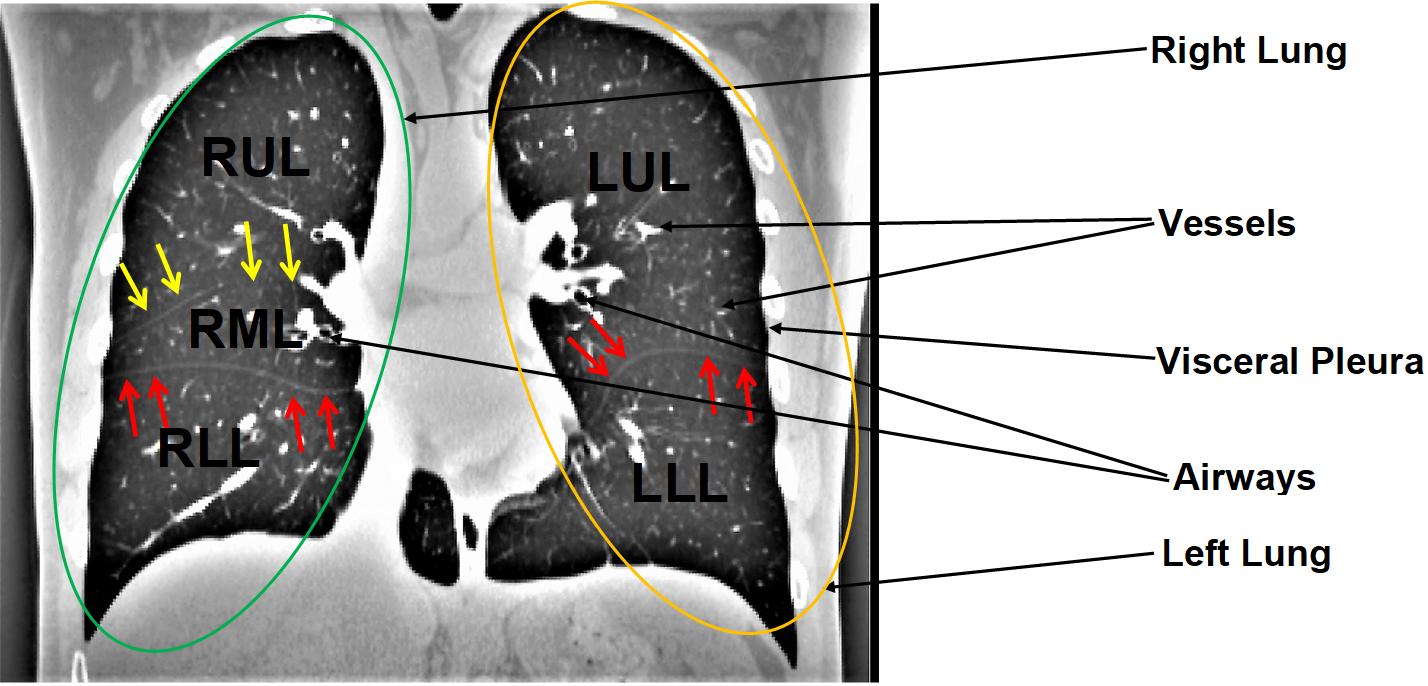} 
\caption{A coronal lung CT slice with visible fissures. Major fissures are denoted by red arrows and yellow arrows denote the minor fissure.} 
\label{fig:coronal-fissures}
\end{figure}

Human lungs are divided into five lobes. The right lung has three lobes, namely, right upper lobe (RUL), right middle lobe (RML), and right lower lobe (RLL), which are separated by a minor and a major fissure, whereas the left lung has two lobes, namely, left upper lobe (LUL) and left lower lobe (LLL), separated by a major fissure. Fig.~\ref{fig:coronal-fissures} shows the five lobes separated by major and minor fissures in a coronal CT slice. Each of the five lobes is functionally independent as they have separate bronchial and vascular systems.

\vspace{4pt}
\noindent Automatic lobe segmentation is important for both clinical and technical purposes. In clinical practice, doctors very often base their assessment of a disease severity and the corresponding treatment plan on the affected lung lobe.  As such, upon encountering a disease or lesion in the lung, radiologists may navigate through the nearby slices to identify the affected lobe, especially when the fissure lines are not clearly visible in the target slice. An automatic lobe segmentation model can therefore shorten the CT reading session by continually informing the radiologists about their location in the lung anatomy. From the technical perspective, accurate lung lobe segmentation can improve several subsequent clinical tasks, including nodule malignancy prediction (cancers mostly occur in the left or right upper lobes), automatic lobe-aware report generation for each nodule, and assessment and quantification of pulmonary diseases, by narrowing down the search space to the lung lobes most-likely to be affected. However, identifying fissures poses a challenge for both human and machine perception. \emph{First}, fissures are most often incomplete, not extending to the lobar boundaries. Several studies in the literature have confirmed the incompleteness of fissures
as a very common phenomenon~\citep{Aziz2004}. \emph{Second}, the visual characteristics of lobar boundaries can change in presence of pathologies. Such morphological changes could also be related to the varying thicknesses, locations, and shapes of the fissures. \emph{Third}, there also exist other fissures in the lungs that can be misinterpreted as the major or minor fissures that separate the lobes (e.g., accessory fissures and azygos fissures).

\vspace{4pt}
\noindent To address the need for accurate and robust lobe segmentation, we have pro-
posed a fully automatic and reliable deep learning solution via progressive dense
V-net (PDV-net). The PDV-net model takes entire CT volume and through three dense feature blocks, generates the segmentation progressively improving at each pathway. Our model generates accurate segmentation of the lung lobes
in about 2 seconds in only a single forward pass of the network, eliminating the
need for any user interactions or any prior segmentation of lungs, vessels, or
airways, which are common assumptions in the design of existing models.

\section{Related Work}
Various automatic and semi-automatic approaches have been proposed for lung lobe segmentation. Despite the methodological differences, the existing approaches are similar in that they require either prior segmentation of airways and vessels (e.g., \citet{Bragman2017}) or demand previously defined atlases (e.g., \citet{vanRikxoort2010} and \citet{Ross2010}). Therefore, they suffer from  slow execution time, cumbersome process of generating the atlas, and relatively lower performance for pathological cases. A significant shift from this common trend is the work of ~\citet{George2017} wherein a 2D fully convolutional neural network followed by a 3D random walker algorithm is used to segment lobes. However, their method still relied on the random walker algorithm whose optimal parameters could change from one dataset to another. It is most desirable to have an end-to-end solution that does not rely on any subsequent heuristic method. 

\vspace{4pt}
\noindent In the presented work, we mitigate the aforementioned limitations, namely reliance on prior masks, slow runtime, and lack of robustness by an end-to-end, single-pass, deep-learning-based framework that does not rely on any prior airway/vessel segmentation, anatomical knowledge, or atlases.

\section{Method}
\begin{figure}[t]
\centering
\includegraphics[width=\linewidth]{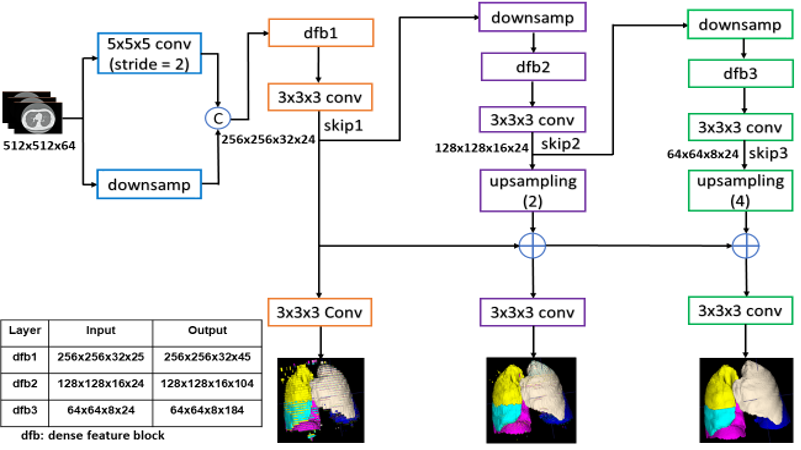}
\caption{PDV-net model for the segmentation of lung lobes. Segmentation outputs at different pathways are progressively improved for the final result.}
\label{fig:dense_vnet}
\end{figure}

We combine the ideas from dense V-network~\cite{Gibson2018} and progressive holistically nested networks \citep{harrison2017} to obtain a new architecture: progressive dense V-network (PDV-net), an end-to-end solution for organ segmentation in 3D volumetric data. Our proposed architecture is illustrated in Fig.~\ref{fig:dense_vnet}. As seen, the input to the network is first down-sampled  and concatenated with a strided $5\times 5\times 5$ convolution of the input with 24 kernels. The concatenation result is then passed to 3 dense feature blocks, each consisting of 5, 10, and 10 densely-wired convolution layers respectively. The growth rates of dense blocks are set to 4, 8, and 16 respectively.  All the convolutional layers in a dense block have a kernel size of $3\times 3\times 3$ and are followed by batch normalization and parametric rectified linear units (PReLU).  

\vspace{4pt}
\noindent Consecutively, the outputs of the dense feature blocks are utilized in low and high resolution passes via convolutional
down-sampling and skip connections. This enables the generation of feature maps at three different resolutions. The outputs of the skip connections of the second and third dense feature blocks are further up-sampled in order to be consistent with the size of the output in the first skip connection. The feature maps from skip1 are passed to a convolutional layer followed by a softmax, which outputs the probability maps. In the second pathway, the feature maps from skip1 and skip2 are merged and the output probability maps are produced by a convolutional layer followed by softmax. Similarly, we get the final segmentation result from the merged feature maps resulted from the skip2 and skip3 connections. Unlike dense V-net, PDV-net generates the final output by progressively improving the outputs at previous pathways. To train the suggested architecture, we choose to use a dice-based loss function~\cite{Milletari2016} at each stage of the progressive architecture.

\section{Experiments}
{\bf Datasets:} We used 3 public datasets to evaluate our models. {\bf First}, we selected a subset of chest CTs (354 cases) from the publicly available LIDC dataset for annotation. To ensure variability in the data, CT scans were selected such that both challenging and visible fissures are well-represented in the dataset. The ground truth masks were generated in a semi-automatic fashion by multiple observers using 3D Slicer. To mitigate bias in the ground truth, the generated masks were later refined and validated by an expert radiologist. The dataset was split into 270 training and 84 test cases. 10\% of the training set was utilized as the validation set.  {\bf Second}, we selected 154 CTs from LTRC database. The LTRC dataset includes lobe masks for pathological cases that have clear evidence of COPD or ILD diseases, including emphysema and fibrosis. The LTRC cases allow us to measure the robustness of our model against pathologies in the lungs. {\bf Third}, we used 55 cases of the Lobe and Lung Analysis (LOLA11) challenge~\citep{LOLA2011} and submitted the results to the challenge organizers for evaluation.  

\vspace{4pt}
\noindent{\bf Baselines for comparison:} We used a U-Net architecture~\citep{Ronneberger2015} and a dense V-Net for comparison. The former is used in the most recent published article \cite{George2017} for lung lobe segmentation and the latter is a strong baseline for comparison, which we use for the first time for lung lobe segmentation.

\vspace{4pt}
\noindent{\bf Implementation Details:} For the proposed model and dense V-Net, the training volumes were first normalized, followed by rescaling to $512\times512\times64$, using 1 NVIDIA Titan XP GPU. Due to the large memory footprint of the model, the gradient check-pointing method~\citep{Bulatov2018} was used for memory-efficient back-propagation. In addition, batch-wise spatial dropout~\cite{Gibson2018} is incorporated for regularization purposes. The training was performed on a Intel(R) Xeon(R) CPU E5-2697 v4@2.30GHz machine. We used the Adam optimizer~\citep{Kingma2014} with an learning rate of 0.01 and a weight decay of $10^{-7}$.

\vspace{4pt}
\noindent For the 2D U-net implementation, we trained the network with axial slices from all the training volumes, each sized $512\times 512$ and normalized to have values between 0 and 1. To avoid over-fitting to the background class, we used only the axial slices, wherein at least one lung lobe is present. We further used the Adam optimizer with a learning rate of $5\times 10^{-5}$ and batches of 10 images.

\begin{table}[t]
\setlength{\tabcolsep}{4pt}
\centering
\caption{Performance comparison of the proposed 3D progressive dense V-net with the 2D U-net and 3D dense V-net models in segmenting 84 LIDC and 154 LTRC cases. Mean Dice score and standard deviation for each lobe have been reported.}
\label{table:lidc-ltrc}
\resizebox{\linewidth}{!}{%
\begin{tabular}{ c | c c c c c c c}
\hline
\hline\noalign{\smallskip}
Dataset & Model & RUL & RML & RLL & LUL & LLL & Overall\\ 
\noalign{\smallskip}
\hline
\hline
\noalign{\smallskip}
\multirow{3}{*}{LIDC(84)} 

    & 2D U-Net &  
0.908 $\pm$ 0.049 & 
0.844 $\pm$ 0.076 & 
0.940 $\pm$ 0.054 & 
0.959 $\pm$ 0.042 & 
0.949 $\pm$ 0.056 & 
0.920 $\pm$ 0.043\\

\noalign{\smallskip}
\cline{2-8}
\noalign{\smallskip}

    & 3D DV-Net & 
0.929 $\pm$ 0.036 & 
0.873 $\pm$ 0.058 &
0.951 $\pm$ 0.018 & 
0.958 $\pm$ 0.020 &
0.949 $\pm$ 0.041 & 
0.932 $\pm$ 0.023 \\ 

\noalign{\smallskip}
\cline{2-8}
\noalign{\smallskip}

    & 3D PDV-Net & 
{\bf 0.937 $\pm$ 0.031} & 
{\bf 0.882 $\pm$ 0.057} & 
{\bf 0.956 $\pm$ 0.017} & 
{\bf 0.966 $\pm$ 0.014} &
{\bf 0.966 $\pm$ 0.037} & 
{\bf 0.939 $\pm$ 0.020} \\ 

\noalign{\smallskip}
\hline
\noalign{\smallskip}

\multirow{3}{*}{LTRC(154)} 
    & 2D U-Net &  
0.914 $\pm$ 0.039 & 
0.866 $\pm$ 0.054 & 
0.952 $\pm$ 0.023 & 
0.961 $\pm$ 0.023 & 
0.954 $\pm$ 0.021 & 
0.929 $\pm$ 0.025\\

\noalign{\smallskip}
\cline{2-8}
\noalign{\smallskip}
  & 3D DV-Net & 
0.949 $\pm$ 0.013 & 
0.901 $\pm$ 0.021 & 
0.959 $\pm$ 0.009 & 
0.961 $\pm$ 0.007 &
0.958 $\pm$ 0.012 &
0.946 $\pm$ 0.008\\ 

\noalign{\smallskip}
\cline{2-8}
\noalign{\smallskip}

  & 3D PDV-Net & 
{\bf 0.952 $\pm$ 0.011} & 
{\bf 0.908 $\pm$ 0.020} & 
{\bf 0.961 $\pm$ 0.008} & 
{\bf 0.966 $\pm$ 0.006} &
{\bf 0.960 $\pm$ 0.010} &
{\bf 0.950 $\pm$ 0.007} \\

\noalign{\smallskip}
\hline
\noalign{\smallskip}

\hline
\end{tabular} }
\end{table}
\setlength{\tabcolsep}{1.4pt}

\begin{figure}
\centering
 \resizebox{0.9\linewidth}{!}{%
  \begin{tabular}{cccccc}
& {\large Slice} & {\large GT} & {\large U-Net} & {\large DV-Net} & {\large PDV-Net}\\
\noalign{\smallskip}
    \rotatebox{90}{\large Left Lung} &
    \includegraphics[width=0.2\linewidth, trim={12.0cm 4.0cm 11.0cm 4.0cm},clip]{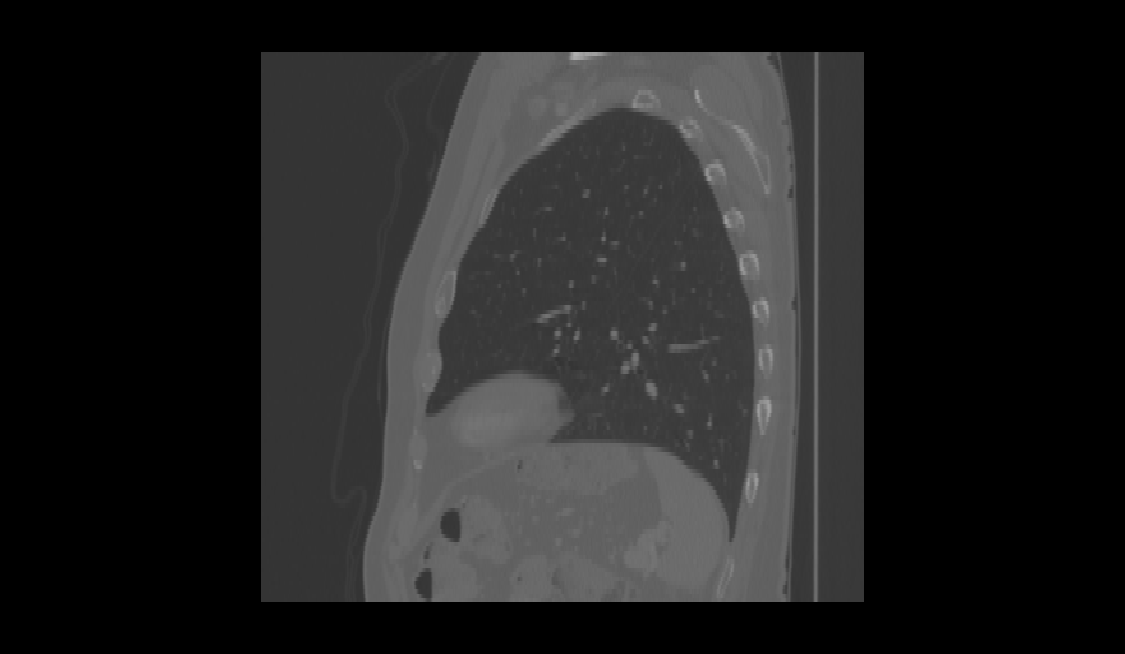} 
    & 
    \includegraphics[width=0.2\linewidth, trim={12.0cm 4.0cm 11.0cm 4.0cm},clip]{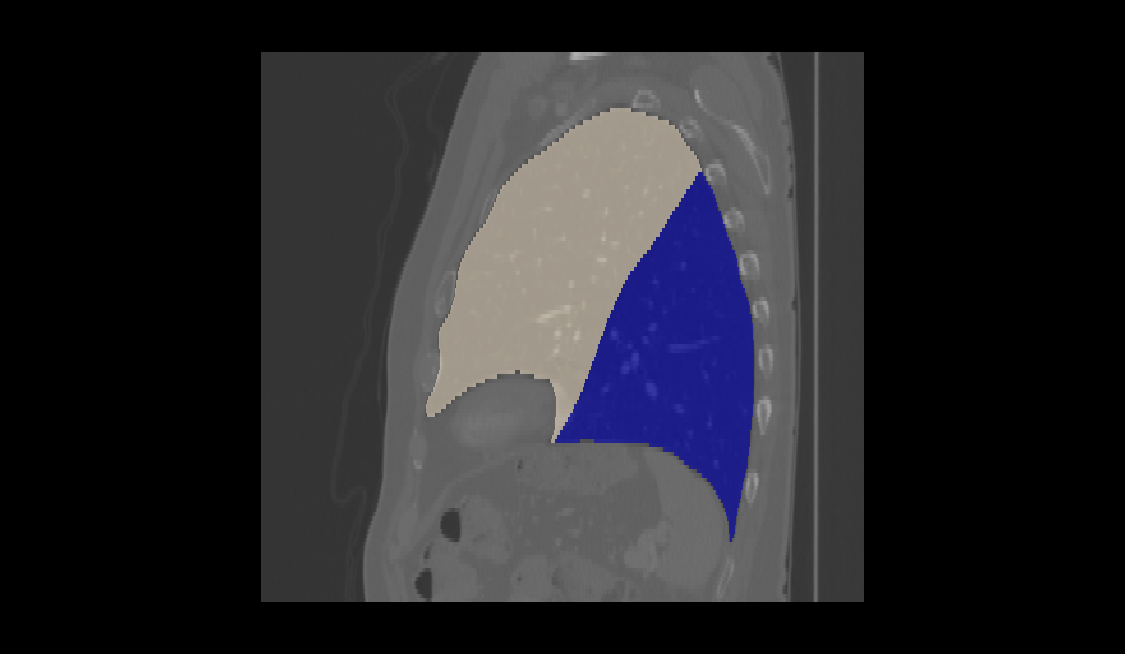} &
    \includegraphics[width=0.2\linewidth, trim={12.0cm 4.0cm 11.0cm 4.0cm},clip]{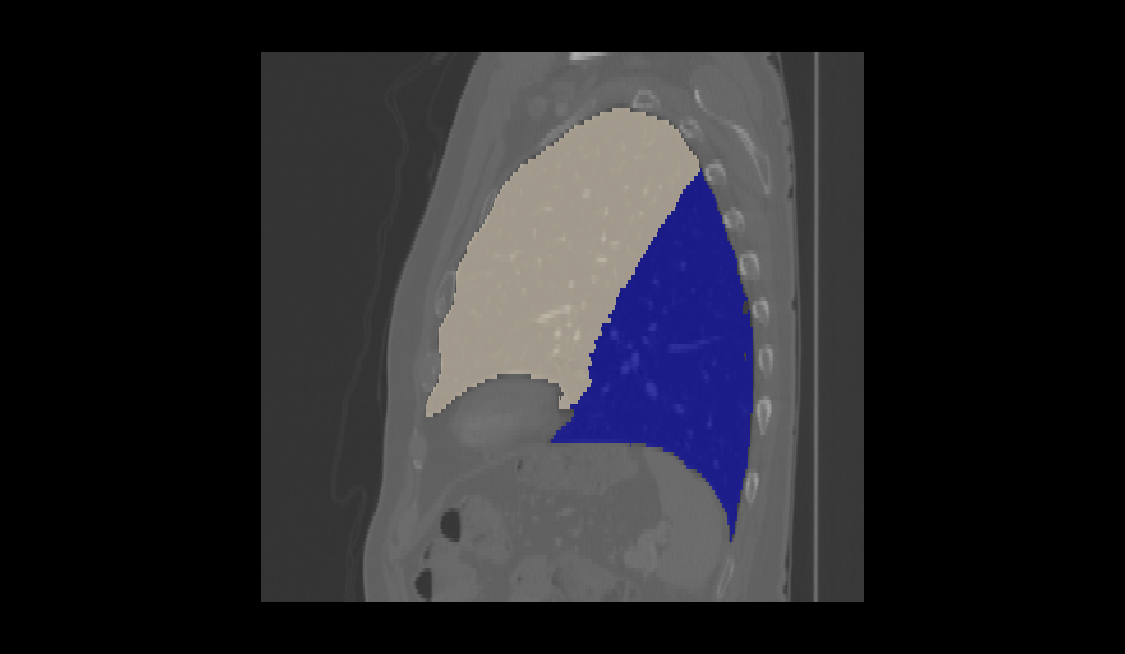} &
    \includegraphics[width=0.2\linewidth, trim={12.0cm 4.0cm 11.0cm 4.0cm},clip]{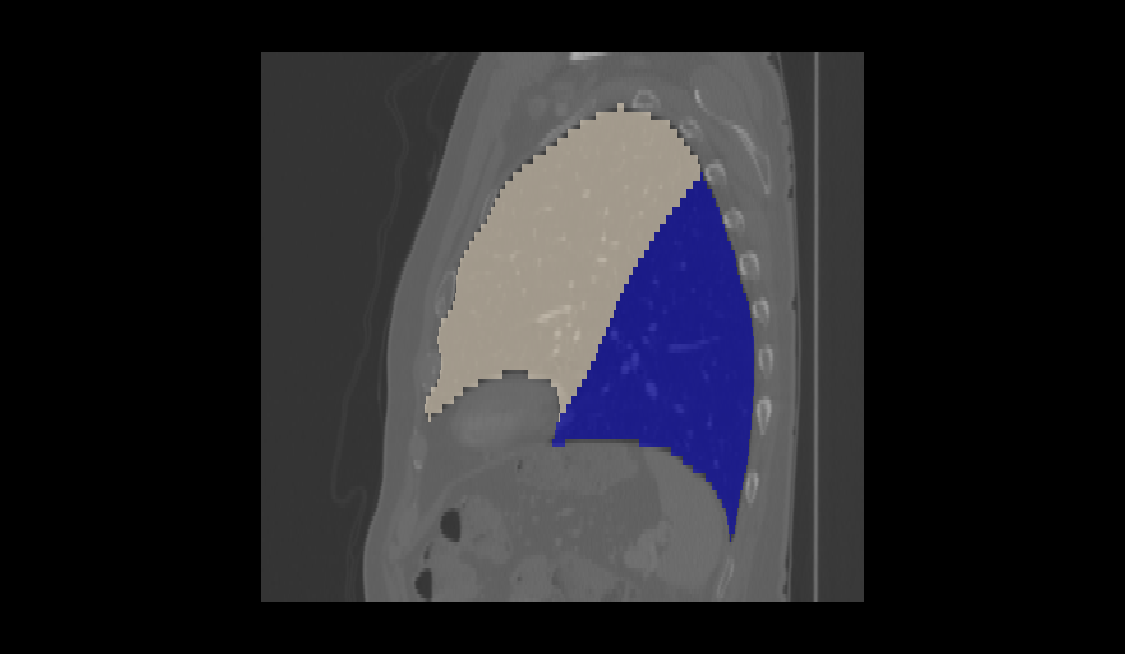} &
    \includegraphics[width=0.2\linewidth, trim={12.0cm 4.0cm 11.0cm 4.0cm},clip]{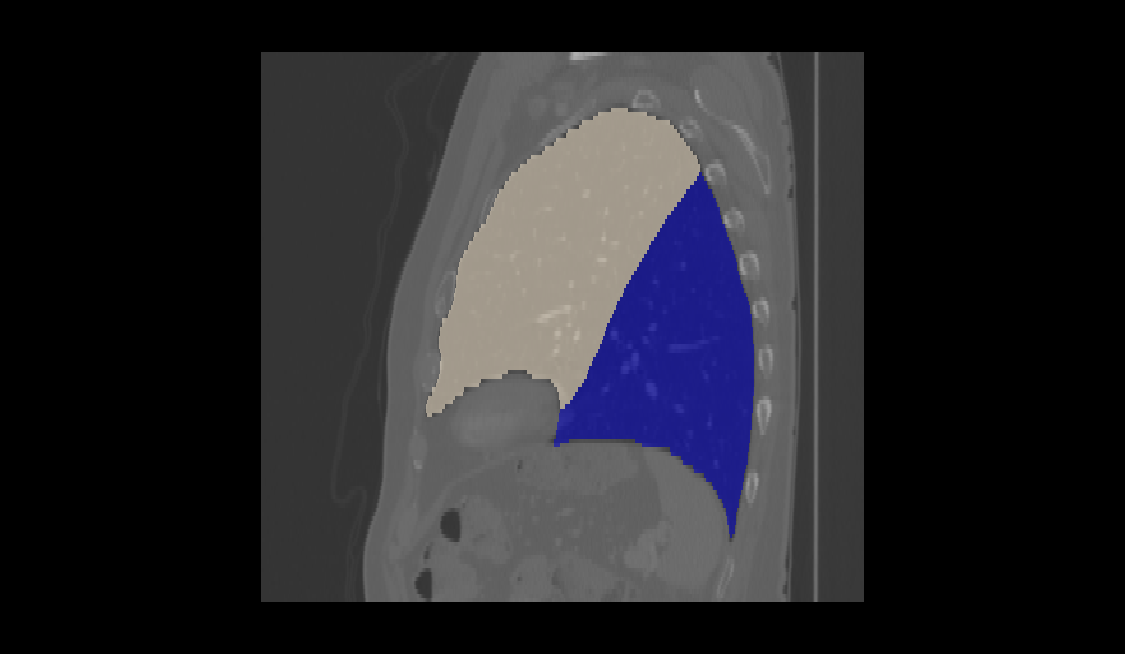}
    \smallskip\\
    \rotatebox{90}{\large Right Lung} &
    \includegraphics[width=0.2\linewidth, trim={12.0cm 4.0cm 11.0cm 4.0cm},clip]{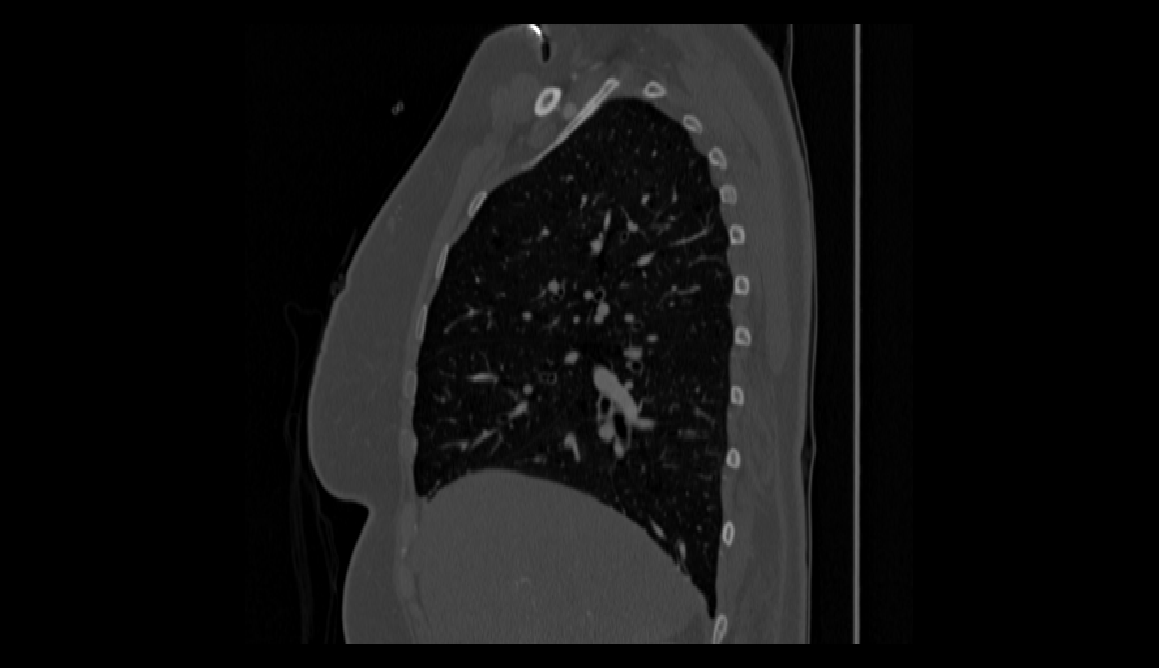} &
	\includegraphics[width=0.2\linewidth, trim={12.0cm 4.0cm 11.0cm 4.0cm},clip]{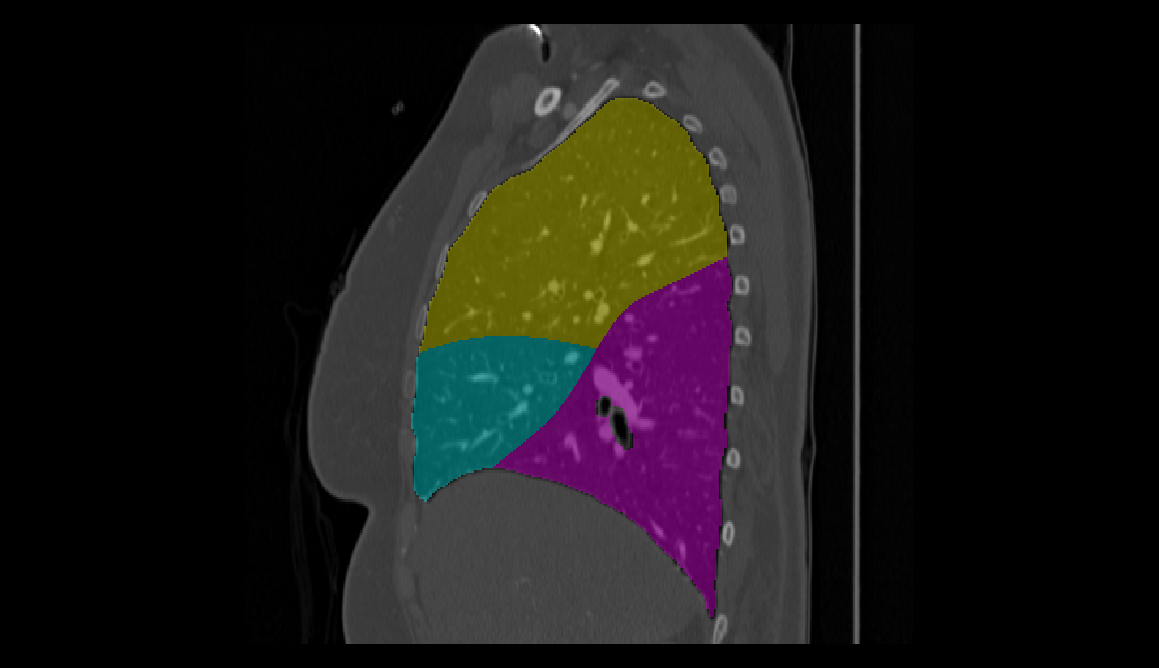} &
    \includegraphics[width=0.2\linewidth, trim={12.0cm 4.0cm 11.0cm 4.0cm},clip]{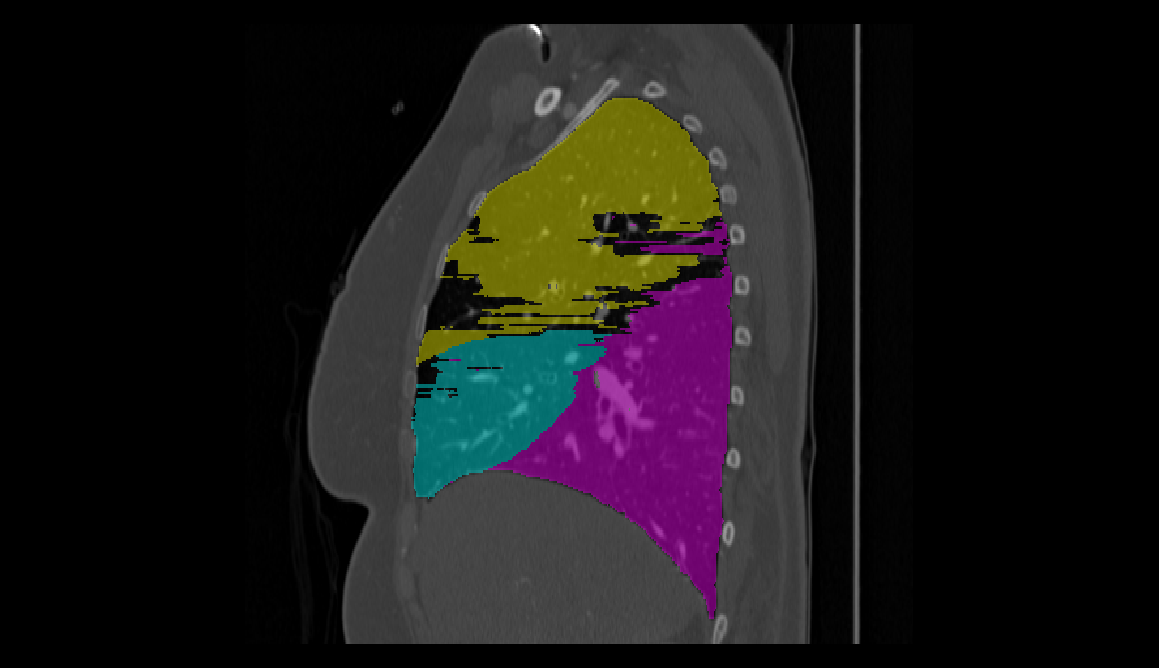} &
    \includegraphics[width=0.2\linewidth, trim={12.0cm 4.0cm 11.0cm 4.0cm},clip]{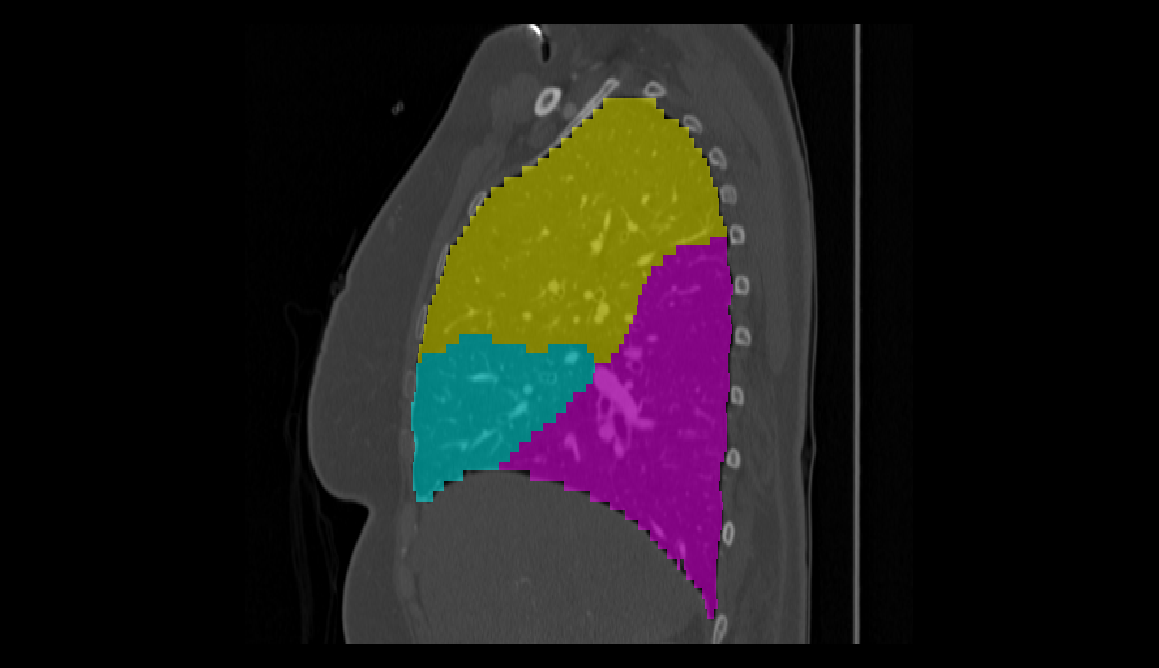} &
    \includegraphics[width=0.2\linewidth, trim={12.0cm 4.0cm 11.0cm 4.0cm},clip]{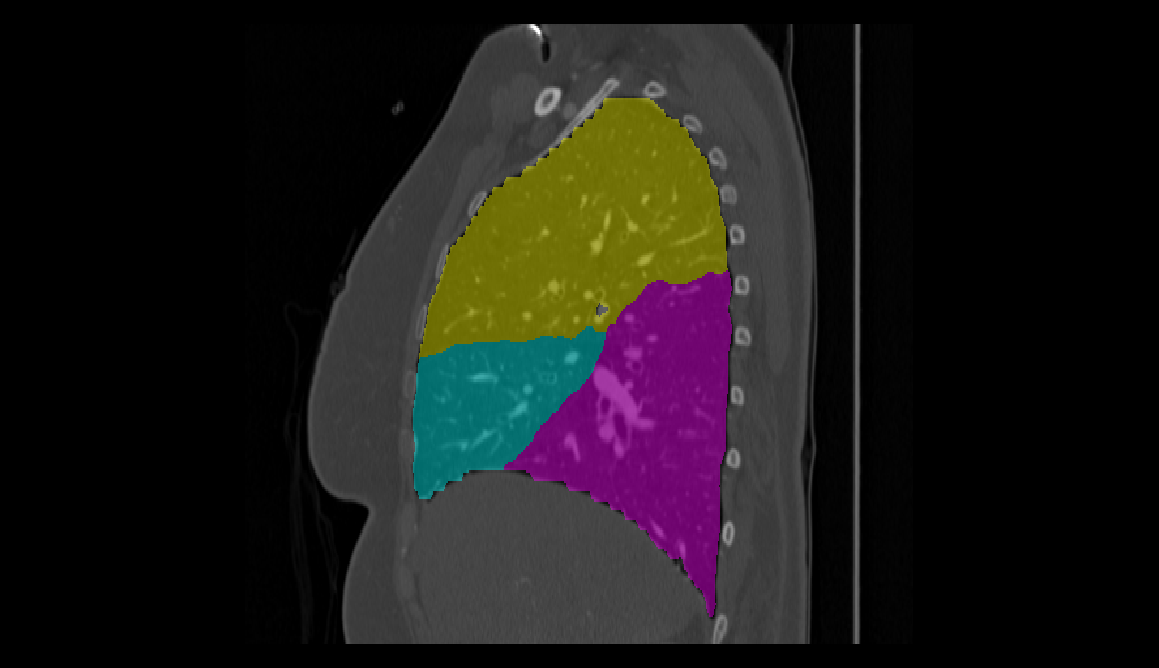}
    \smallskip\\
    &
    \rotatebox{0}{\large 3D} &
	\includegraphics[width=0.2\linewidth, trim={16.0cm 5.0cm 11.0cm 5.0cm},clip]{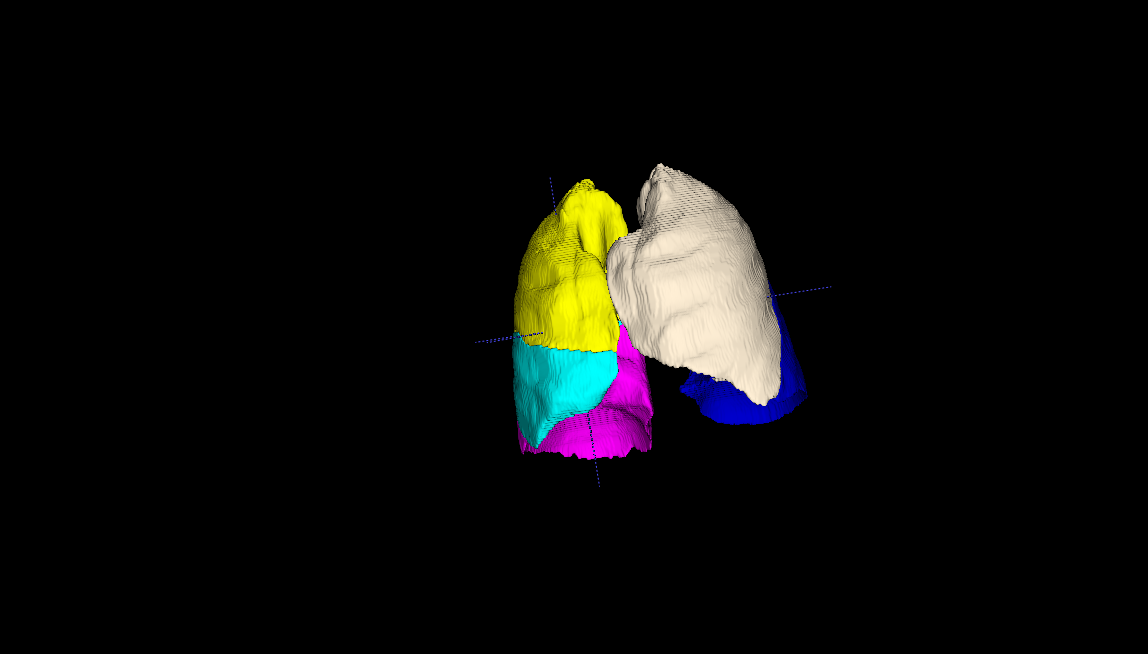} &
     \includegraphics[width=0.2\linewidth, trim={16.0cm 5.0cm 11.0cm 5.0cm},clip]{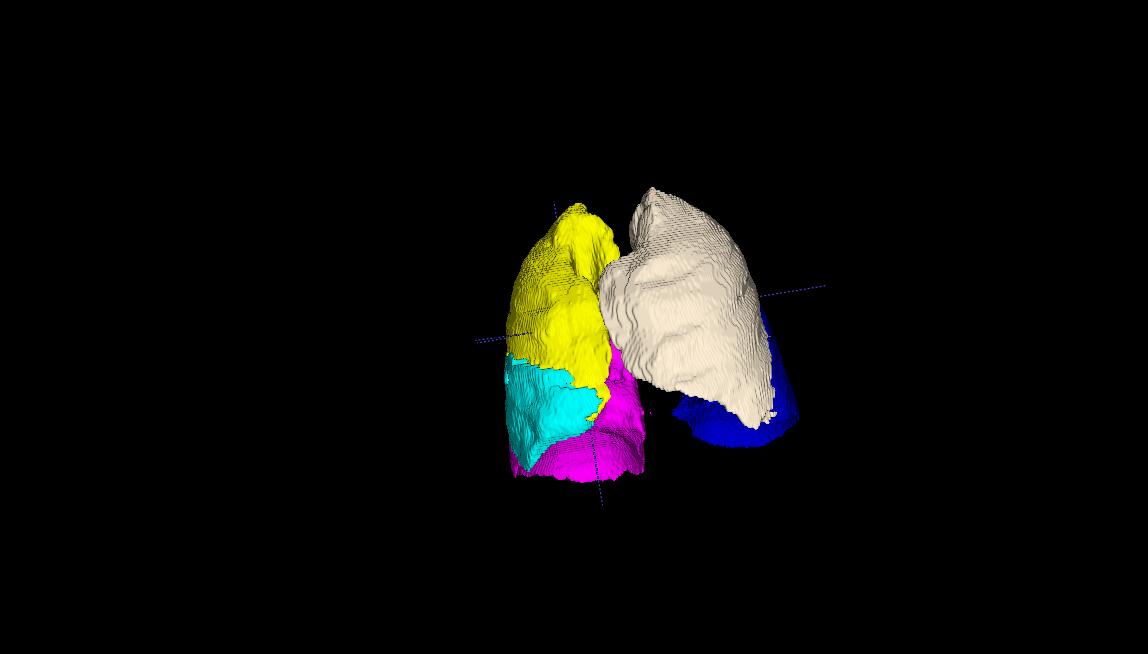} &
    \includegraphics[width=0.2\linewidth, trim={16.0cm 5.0cm 11.0cm 5.0cm},clip]{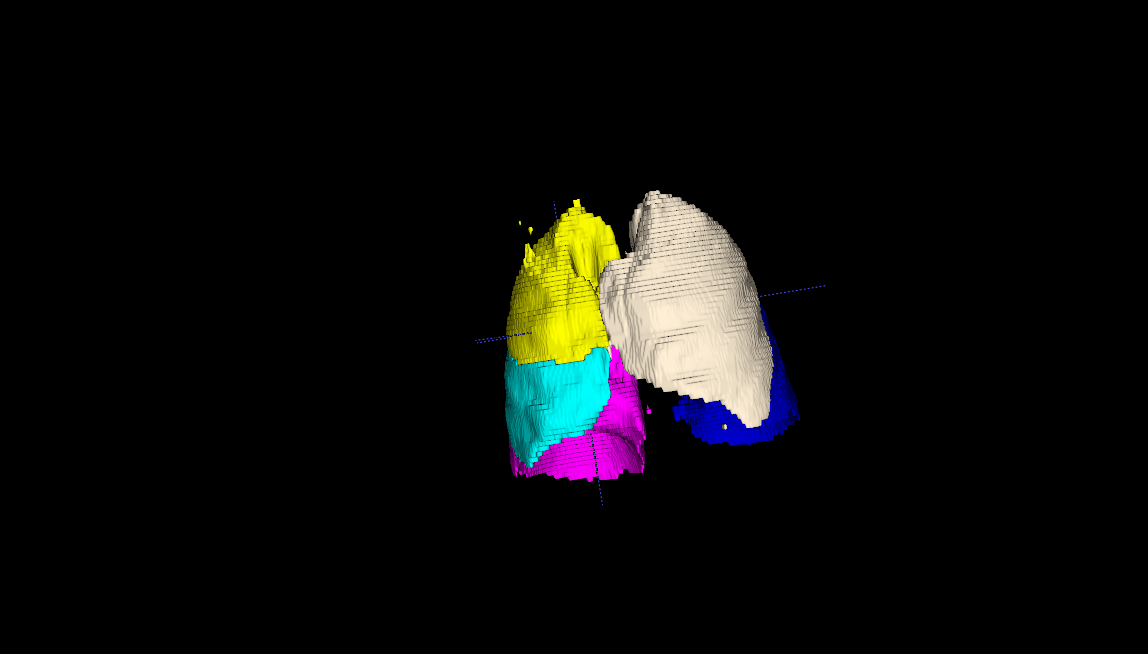} &
    \includegraphics[width=0.2\linewidth, trim={16.0cm 5.0cm 11.0cm 5.0cm},clip]{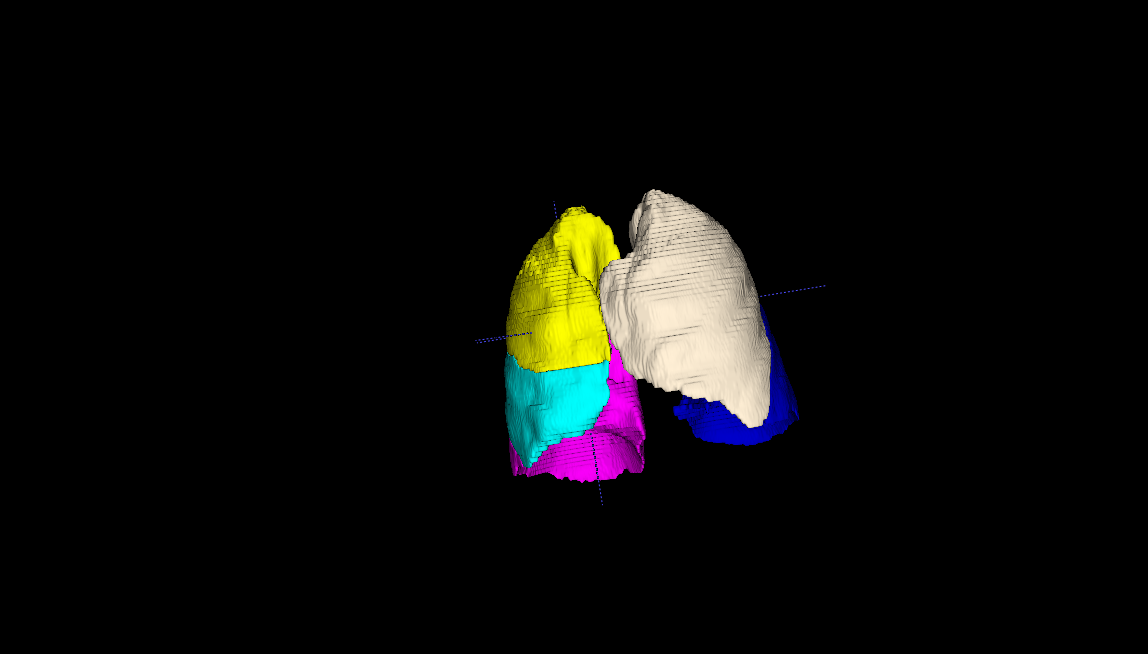} 
    
 \end{tabular} }
  \caption{Qualitative comparison between the proposed 3D progressive dense V-Net (PDV-Net), dense V-Net (DV-Net), and U-net. Note how noisy patches are removed from the final segmentation generated by PDV-Net. Color coding: almond:~LUL, blue:~LLL, yellow:~RUL, cyan:~RML, pink:~RLL.}
  \label{fig:vis-lidc}
\end{figure}

\vspace{4pt}
\noindent{\bf LIDC Results:} Table~\ref{table:lidc-ltrc} shows the calculated overall and lobe-wise Dice scores for each of the models. The proposed progressive dense V-net model, with an overall score of $0.939 \pm 0.020$, significantly outperformed the 2D model, with an overall score of $0.9201 \pm 0.0431$. As is evident in Table~\ref{table:lidc-ltrc}, the 3D progressive dense V-net yields consistently larger Dice score for each of the lung lobes against both dense V-net and U-net. Moreover, the lower standard deviation for each lobe indicates that the progressive model is more robust. We have also shown a qualitative comparison between the 3 models in Fig.~\ref{fig:vis-lidc} where the lung fissures are better captured by our progressive dense V-net model than by 2D U-net and dense V-net.  

\begin{figure}[t]
\centering
\includegraphics[width=0.32\linewidth]{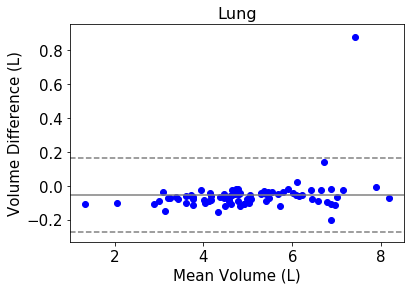}
\hfill
\includegraphics[width=0.32\linewidth]{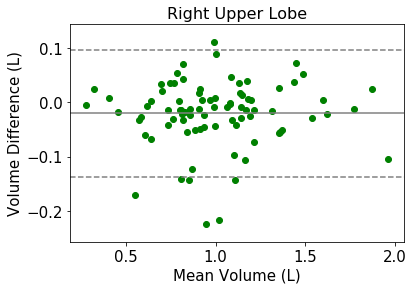}
\hfill
\includegraphics[width=0.32\linewidth]{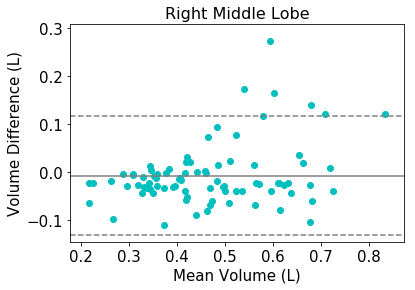} \\

\includegraphics[width=0.32\linewidth]{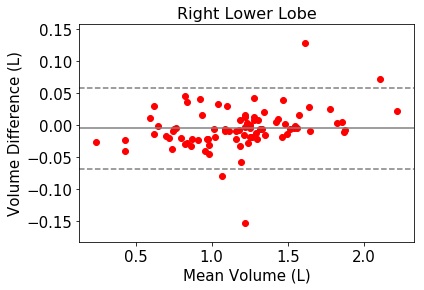}
\hfill
\includegraphics[width=0.32\linewidth]{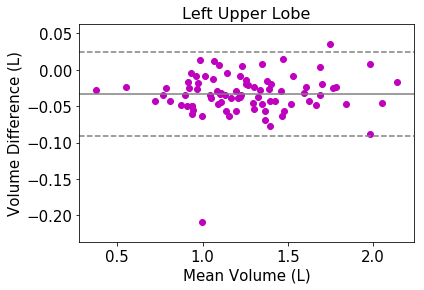}
\hfill
\includegraphics[width=0.32\linewidth]{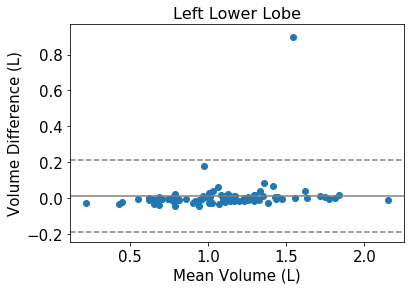}
\caption{Bland-Altman plots show the agreement between 3D progressive dense V-net and ground truth.}
\label{fig:bland-altman}
\end{figure}

\vspace{4pt}
\noindent We further used Bland-Altman plots to measure the agreement between our progressive dense V-net and ground truth segmentations of the 84 LIDC cases (Fig.~\ref{fig:bland-altman}). A good agreement was observed between our segmentation model and ground truth in every plot (Lung and LLL being the two best agreements). Pearson correlation showed that all six volume sets in ground truth are strongly correlated with the corresponding six volume sets in the PDV-net segmentation, with $p < 0.001$. 

\vspace{4pt}
\noindent{\bf LTRC Results:} Table~\ref{table:lidc-ltrc} shows that the 3D progressive dense V-net achieves an average Dice score of $0.950 \pm 0.007$, significantly improving the dense V-net ($0.946 \pm 0.008$). Once again, the progressive dense V-net model outperformed the 2D U-net model with an average Dice score of $0.929 \pm 0.025$. Individual lobes were segmented better in the proposed 3D progressive dense V-net model than in the 3D dense V-net and the 2D U-net models (Table \ref{table:lidc-ltrc}). Note that the LTRC dataset includes many pathological cases where the fissure lines are either invisible, distorted, or absent in presence of pathologies such as emphysema, fibrosis, etc. As a result, lobe segmentation becomes more challenging. Nevertheless, our model performed well in segmenting lobes in pathological cases from the LTRC dataset. Moreover, our model outperformed the model of~\citet{George2017} in segmenting the LTRC cases both in Dice score (0.941 $\pm$ 0.255) and inference speed (4-8 minutes per case).

\vspace{4pt}
\noindent{\bf LOLA11 Results:} The segmentation results on the LOLA11 cases submitted online were evaluated as overlap (Jaccard) scores. To be consistent with our previous analyses, we converted the Jaccard scores to Dice scores. The results are shown in Table~\ref{table:lola-score}. Our method achieved an overall Dice score of 0.934, which is very competitive with the state-of-the-art \citep{Bragman2017} with a Dice score of 0.938, while outperforming the methods of~\citet{Giuliani2018} and~\citet{vanRikxoort2010}.

\begin{table}[t]
\setlength{\tabcolsep}{4pt} \centering
\caption{Performance evaluation of 3D PDV-Net models on 55 LOLA cases:
showing lobe-wise mean Dice scores, standard deviations, median
scores, first quartiles, and third quartiles}
\label{table:lola-score}
\resizebox{0.7\linewidth}{!}{%
\begin{tabular}{ c c c c c }
\hline
\hline\noalign{\smallskip}
Lobe & Mean $\pm$ SD & $Q_1$ & Median & $Q_3$\\ 
\noalign{\smallskip}
\hline
\hline
\noalign{\smallskip}
RUL & 0.9518 $\pm$ 0.1750 & 0.9371 & 0.9688 & 0.9881 \\
\noalign{\smallskip}
\hline
\noalign{\smallskip}
RML & 0.8621 $\pm$ 0.4149 & 0.8107 & 0.9284 & 0.9663 \\ 
\noalign{\smallskip}
\hline
\noalign{\smallskip} 
RLL & 0.9581 $\pm$ 0.1993 & 0.9621 & 0.9829 & 0.9881 \\
\noalign{\smallskip}
\hline
\noalign{\smallskip} 
LUL & 0.9551 $\pm$ 0.2160 & 0.9644 & 0.9834 & 0.9924 \\
\noalign{\smallskip}
\hline
\noalign{\smallskip} 
LLL & 0.9342 $\pm$ 0.3733 & 0.9546 & 0.9805 & 0.9902 \\
\noalign{\smallskip}
\hline
\hline
\noalign{\smallskip}
Overall & 0.9345 \\
\citep{Giuliani2018} & 0.9282 \\
\citep{Bragman2017} & 0.9384 \\
\citep{vanRikxoort2010} & 0.9195 \\
\noalign{\smallskip}
\hline
\hline
\multicolumn{5}{|c|}\footnote*{\small {Jaccard score to Dice score conversion: $\mathrm{Dice} = 2\times \mathrm{Jaccard}/(1+\mathrm{Jaccard})$}}
\end{tabular}}

\end{table}
\setlength{\tabcolsep}{1.4pt}

\vspace{4pt}
\noindent{\bf Robustness Analysis:} We further investigated the robustness of our model by grouping the 84 LIDC cases in three ways. For the first grouping, the Dice scores were put in three different Z-spacing buckets: $\hbox{Z-spacing}\le 1$, $1<\hbox{Z-spacing}<2$, and $\hbox{Z-spacing}\ge 2$. In the second grouping, the Dice scores were put in four manufacturer buckets: GE, Philips, Siemens, and Toshiba. In the third grouping, the Dice scores were grouped according to the reconstruction kernel into 3 buckets: soft, lung, and bone.  The one-way ANOVA analysis confirmed that there were no significant differences between the average Dice scores of the buckets within each grouping, suggesting that our model is robust against the choice of reconstruction kernel, size of reconstruction interval, and different CT scan vendors. Moreover, nodule volume in each of the 84 cases does not affect the lobe segmentation performance. There is no correlation between nodule volume and lobe segmentation accuracy, found from Pearson correlation. 

\vspace{4pt}
\noindent We also studied how the segmentation correlation is affected by lung pathologies. For this purpose, we analyzed the correlation between Dice scores and emphysema index (proportion of the lung affected by emphysema) in LTRC cases. According to the Pearson correlation, it was found that lobe segmentation accuracy is not correlated with emphysema index, indicating the robustness of our proposed model in segmenting lobes from pathological cases.  

\vspace{4pt}
\noindent{\bf Speed Analysis:} The proposed 3D progressive dense V-net model takes approximately 2 seconds to segment lung lobes from one CT scan using 1 Nvidia Titan XP GPU, which is six times faster than the 2D U-net model. As per our knowledge from the lung lobe segmentation models available in literature, this is by far the fastest model. Note that no prior published research has yet considered a 3D convolutional model for lung lobe segmentation. 

\section{Conclusions}

Automatic and reliable lung lobe segmentation is a challenging task in the presence of chest pathologies and in the absence of visible, complete fissures. In this paper, we introduced a new 3D segmentation approach, namely, progressive dense V-networks for the automatic, fast, and reliable segmentation of lung lobes from chest CT scans, without any prior segmentation. We evaluated our method using 3 test datasets: 84 cases from LIDC, 154 cases from LTRC, and 55 cases from LOLA11. Our results demonstrated that the suggested model outperforms, or at worst performs comparably to, the state-of-the-art while running at an average speed of 2 seconds per case. Our analyses further demonstrated the robustness of the suggested method against varying configurations of CT reconstruction, choice of CT vendor, and presence of lung pathologies.

\bibliography{dlmia18}

\end{document}


\title{Automatic
Segmentation of Pulmonary Lobes Using a Progressive Dense V-Network\\
(Supplemental Document)}
\maketitle              

\appendix
\section{Base-Line Comparison: 2D U-Net}
In order to construct a baseline model for comparison, we implemented a 2D U-net~\citep{Ronneberger2015} as the only representative of the state-of-the-art. The implemented architecture is symmetric and consists of four contracting and expanding layers, starting with 16 features in the first layer and doubling the number of features in each step. Each contracting layer consists of two $3\times3$ convolutions and a ReLU activation followed by a $2\times2$ max-pooling layer. The expansion path consists of an up-convolution with feature concatenation from the respective contracting layer, and two $3\times3$ convolutions. In addition, all the ReLU layers are preceded by a batch-normalization layer. To improve the training process, we also use a generalized Dice score as the loss function such that the contribution of each class in the image to the gradients is balanced.

\subsection*{LOLA11 Visualization}

\begin{table}
\setlength{\tabcolsep}{4pt}
\centering
\caption{Performance evaluation of 3D progressive dense V-net models on 55 LOLA cases: showing lobe-wise mean Dice scores, standard deviations, median scores, first quartiles, and third quartiles}
\label{table:lola-score}
\resizebox{0.6\columnwidth}{!}{%
\begin{tabular}{ c c c c c }
\hline
\hline\noalign{\smallskip}
Lobe & Mean $\pm$ SD & $Q_1$ & Median & $Q_3$\\ 
\noalign{\smallskip}
\hline
\hline
\noalign{\smallskip}
RUL & 0.9518 $\pm$ 0.1750 & 0.9371 & 0.9688 & 0.9881 \\
\noalign{\smallskip}
\hline
\noalign{\smallskip}
RML & 0.8621 $\pm$ 0.4149 & 0.8107 & 0.9284 & 0.9663 \\ 
\noalign{\smallskip}
\hline
\noalign{\smallskip} 
RLL & 0.9581 $\pm$ 0.1993 & 0.9621 & 0.9829 & 0.9881 \\
\noalign{\smallskip}
\hline
\noalign{\smallskip} 
LUL & 0.9551 $\pm$ 0.2160 & 0.9644 & 0.9834 & 0.9924 \\
\noalign{\smallskip}
\hline
\noalign{\smallskip} 
LLL & 0.9342 $\pm$ 0.3733 & 0.9546 & 0.9805 & 0.9902 \\
\noalign{\smallskip}
\hline
\hline
\noalign{\smallskip}
Overall & 0.9345 \\
~\citep{Giuliani2018} & 0.9282 \\
~\citep{Bragman2017} & 0.9384 \\
~\citep{vanRikxoort2010} & 0.9195 \\
\noalign{\smallskip}
\hline
\hline
\multicolumn{5}{|c|}\footnote*{\small {Jaccard score to Dice score conversion: $\mathrm{Dice} = 2\times \mathrm{Jaccard}/(1+\mathrm{Jaccard})$}}
\end{tabular}}

\end{table}
\setlength{\tabcolsep}{1.4pt}

\begin{figure}
\centering
\subfigure[]{
    \includegraphics[width=0.2\linewidth,trim={12cm 3cm 12cm 1cm},clip]{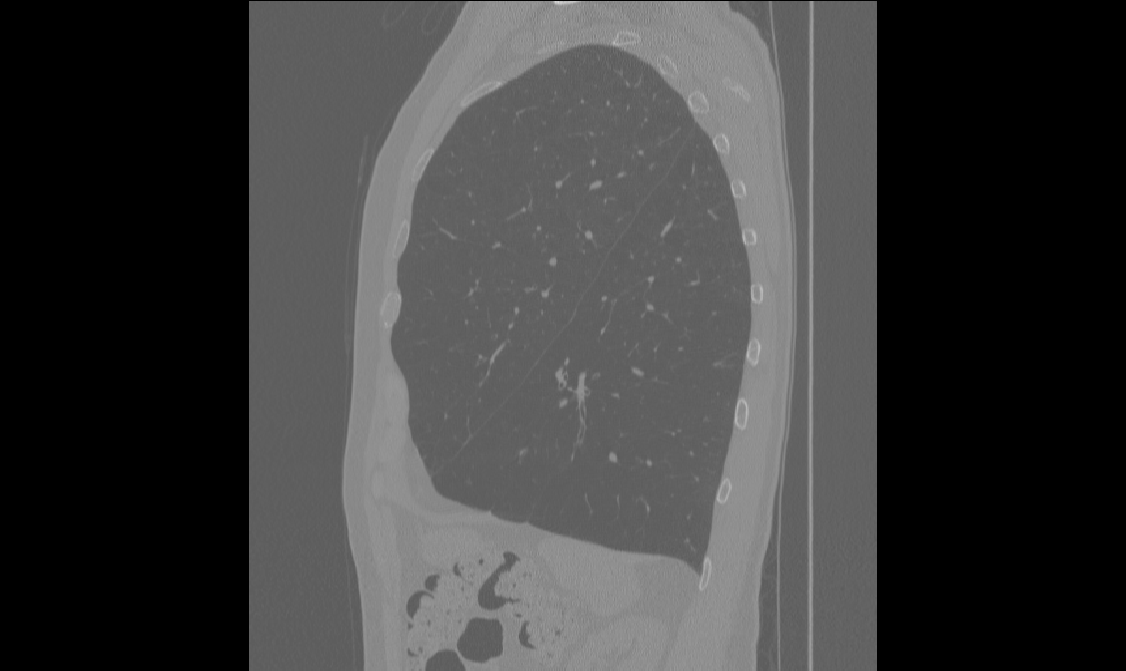} 
    \label{fig:left8-slice}
        }
\subfigure[] {
    \includegraphics[width=0.2\linewidth,trim={12cm 3cm 12cm 1cm},clip]{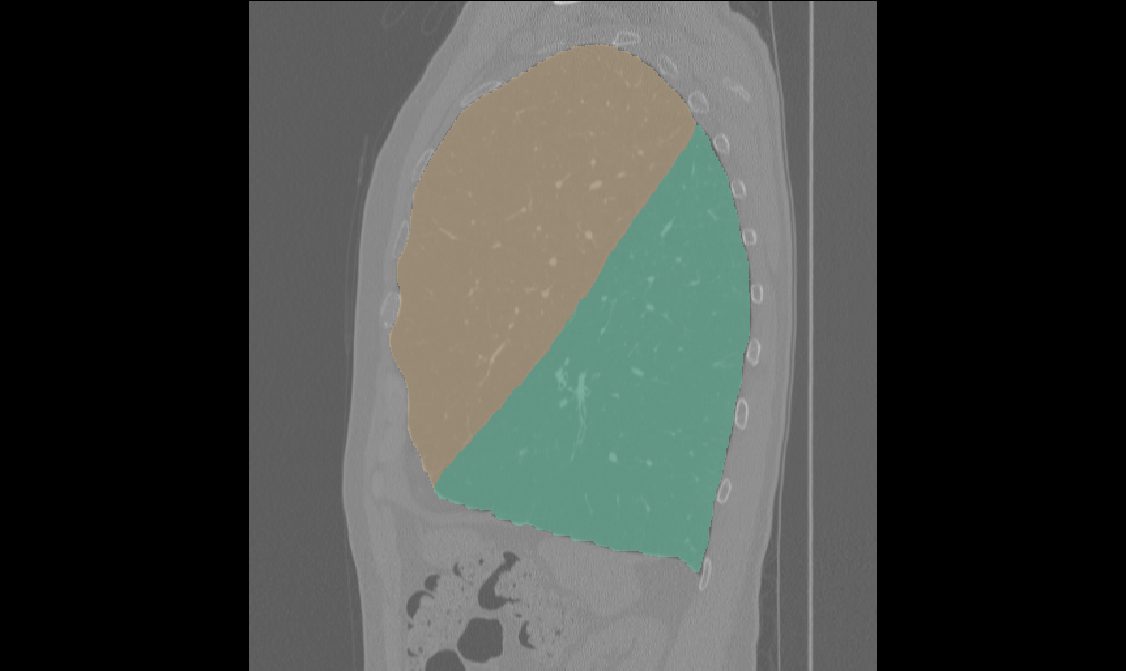} 
    \label{fig:leftt8-seg}
    }
\subfigure[]{
\includegraphics[width=0.2\linewidth,trim={12cm 3cm 12cm 1cm},clip]{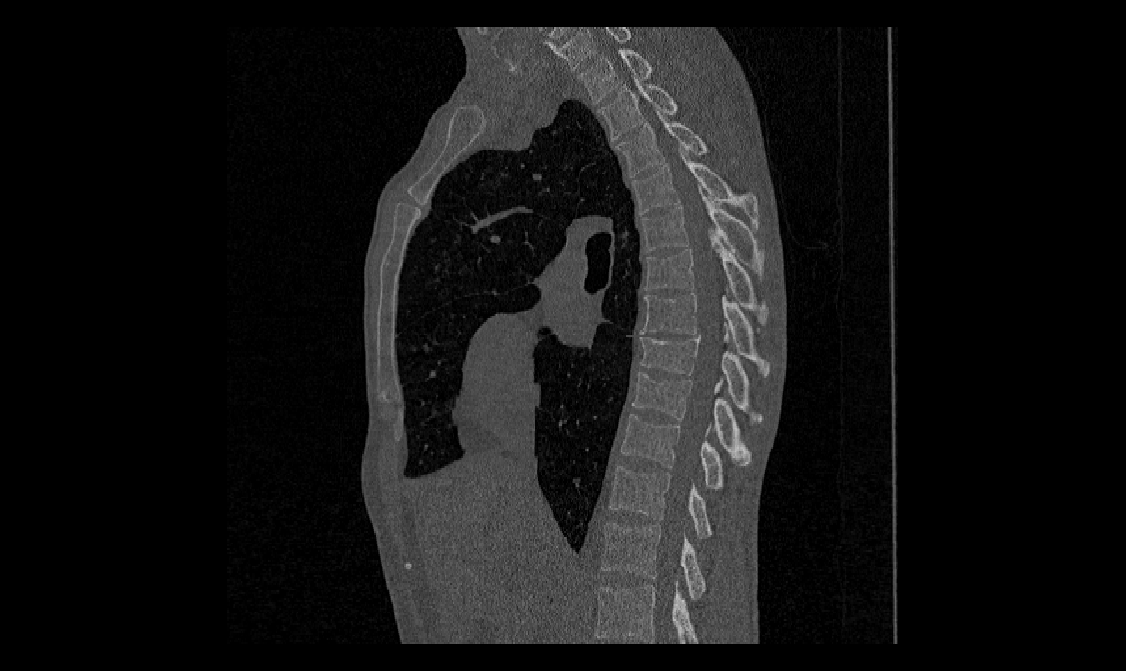} 
\label{fig:case6-slice}
    }
\subfigure[]{
\includegraphics[width=0.2\linewidth,trim={12cm 3cm 12cm 1cm},clip]{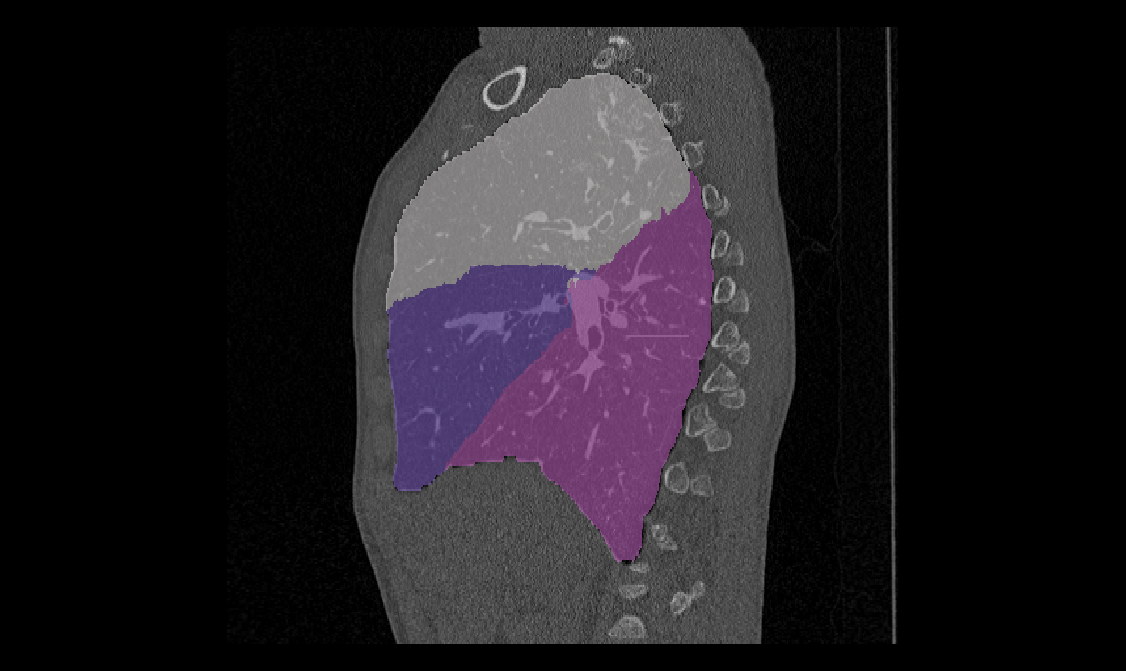}
\label{fig:case6-seg}
}
\subfigure[]{
\includegraphics[width=0.2\linewidth,trim={12cm 2cm 11cm 1cm},clip]{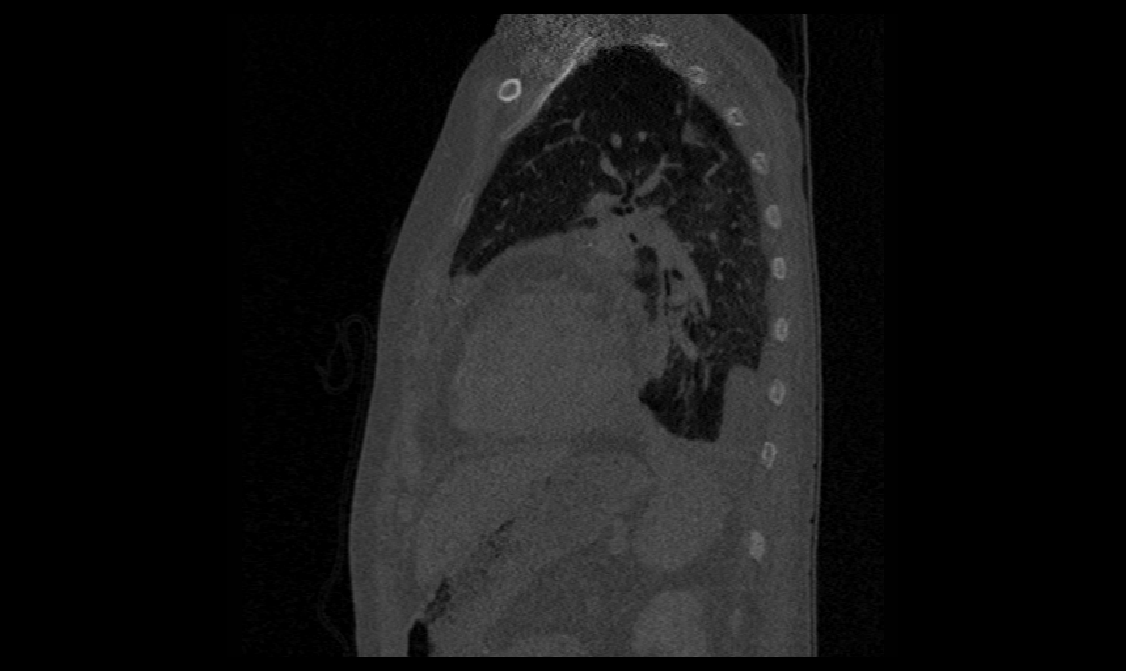}
\label{fig:case21-slice}
    }
\subfigure[]{
\includegraphics[width=0.2\linewidth,trim={12cm 2cm 11cm 1cm},clip]{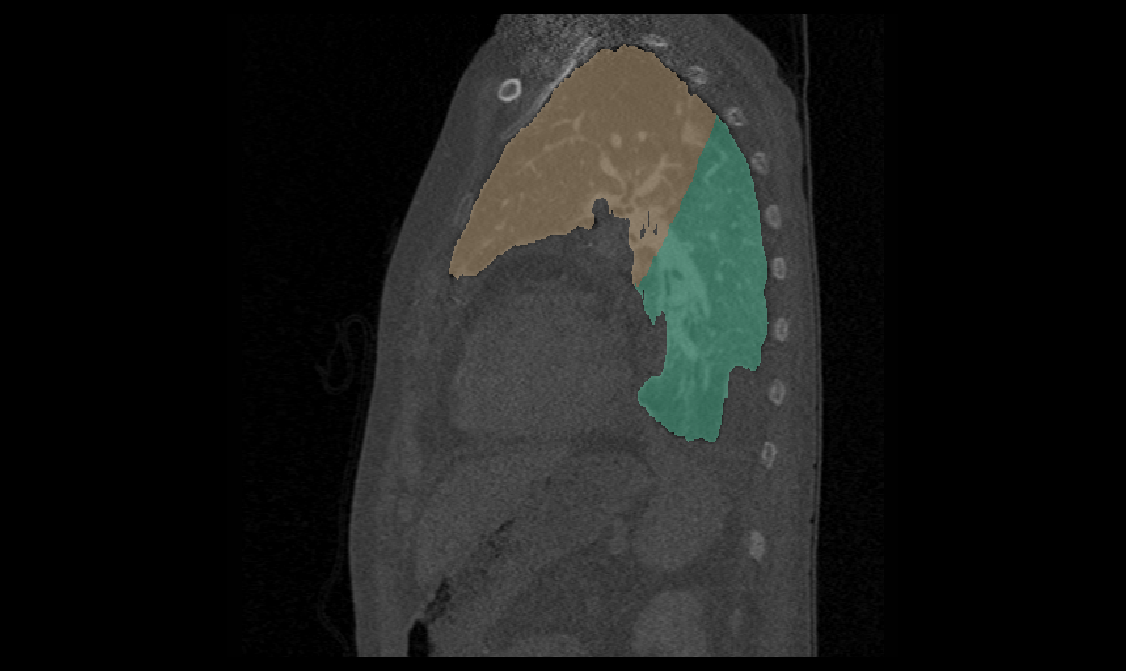} 
\label{fig:case21-seg}
}
\subfigure[]{   
\includegraphics[width=0.2\linewidth,trim={12cm 2cm 11cm 1cm},clip]{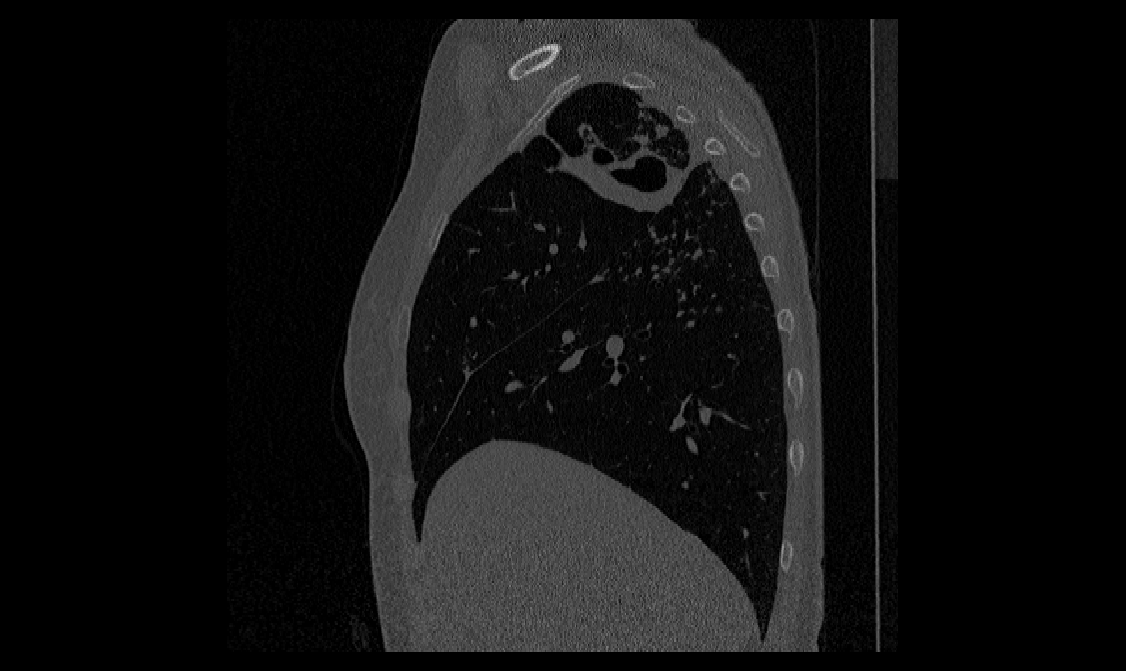}
\label{fig:case55-slice}
}
\subfigure[]{
\includegraphics[width=0.2\linewidth,trim={12cm 2cm 11cm 1cm},clip]{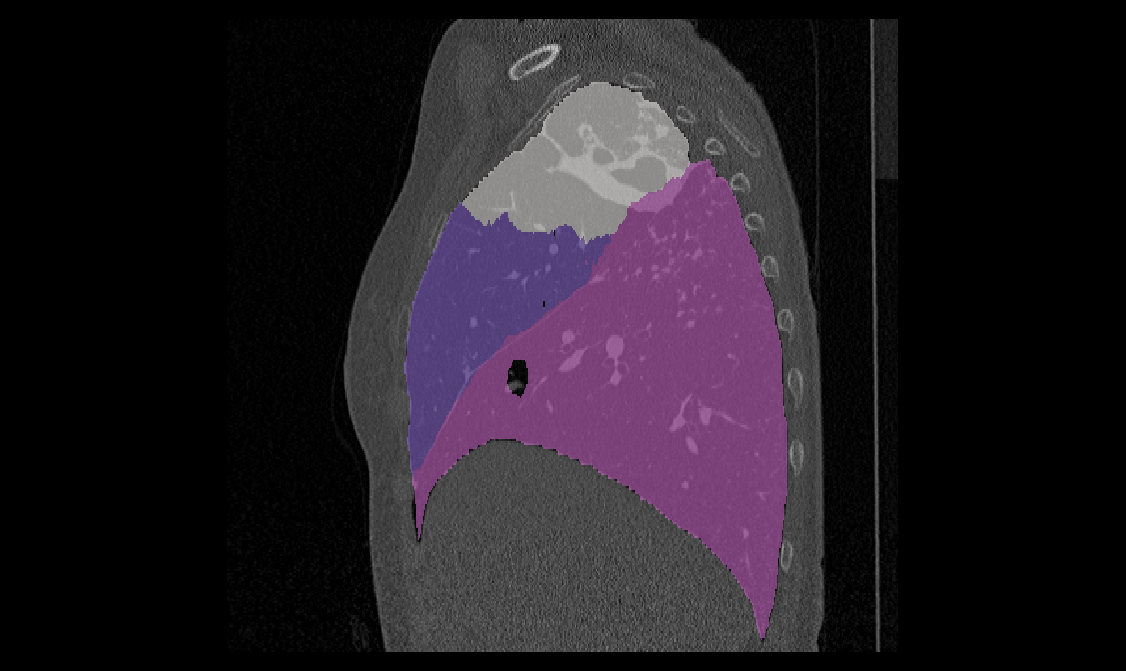}
\label{fig:case55-seg}
}
\caption{Sagittal plane visualization of LOLA11 segmentation by 3D progressive dense V-net: (a) Case 8-slice (left lung), (b) Case 8-segmentation, (c) Case 6-slice (right lung), (d) Case 6-segmentation, (e) Case 21-slice (left lung), (f) Case 21 segmentation, (g) Case 55-slice (right lung), (h) Case 55-segmentation }
  \label{fig:lola-cases}
\end{figure}

Visualizations of LOLA11 cases (Fig.~\ref{fig:lola-cases}) show the segmented slices for some of the cases. For the left lung in case 8 (Fig.~\ref{fig:left8-slice} and Fig.~\ref{fig:leftt8-seg}), the LUL and LLL Dice scores were 0.9940 and 0.9926, respectively. For the right lung in case 6 (Fig.~\ref{fig:case6-slice} and Fig.~\ref{fig:case6-seg}), the scores are as follows: RUL: 0.9580, RML: 0.9480, and RLL: 0.9869. Again, the left lung of case 21 (Fig.~\ref{fig:case21-slice} and Fig.~\ref{fig:case21-seg}), where the segmentation Dice scores were relatively low. For the left lung in case 21, the LUL score was 0.8170 and LLL score was 0.3035. For the right lung in case 55 (Fig.~\ref{fig:case55-slice} and Fg.~\ref{fig:case55-seg}), although the right lower lobe was segmented with a high Dice score of 0.9818, because of the invisibility of the horizontal fissure, the RUL and RML had low segmentation Dice scores of 0.6827 and 0.7499, respectively. 

